\documentclass[conference]{IEEEtran}
\IEEEoverridecommandlockouts
\usepackage{cite}
\usepackage{amsmath,amssymb,amsfonts}
\usepackage{algorithmic}
\usepackage{graphicx}
\usepackage{textcomp}
\usepackage{xcolor}
\def\BibTeX{{\rm B\kern-.05em{\sc i\kern-.025em b}\kern-.08em
    T\kern-.1667em\lower.7ex\hbox{E}\kern-.125emX}}
\begin{document}

\title{Emotional Distraction for Children Anxiety Reduction During Vaccination}

\author{\IEEEauthorblockN{Martina Ruocco}
\IEEEauthorblockA{
\textit{University of Manchester, UK}\\
martina.ruocco@postgrad.manchester.ac.uk}
\and
\IEEEauthorblockN{Marwa Larafa}
\IEEEauthorblockA{
\textit{University of Perugia, Italy}\\}
\and
\IEEEauthorblockN{Silvia Rossi}
\IEEEauthorblockA{
\textit{University of Naples Federico II, Italy}\\
silvia.rossi@unina.it}
}

\maketitle

\begin{abstract}
Social assistive robots are starting to be widely used in pediatric health-care environments with the aim of distracting and entertaining children, and so of reducing a possible state of anxiety. In this paper, we present some initial results of a study (N=69) conducted in a Health-Vaccines Center, where the distraction role of a social robot, which interacts with a child showing an emotional behavior, is compared with the same not showing any emotional social cue. Outcome criteria for the evaluation of the intervention included the parents reported level of anxiety before, during and after the procedure. %, and an external evaluation of the pain.  
\end{abstract}


\section{Introduction}
Social Assistive Robots (SARs) are becoming increasingly used in pediatric health-care where the issue is not only to attract the children attention, but to remain compelling during the interaction \cite{Ahmad2017}. This is even more challenging in the case of an incoming medical procedure when it could be difficult to even get the children attention in the first place. Effective interaction strategies and the ability to keep the engagement during the interaction, in these cases, is extremely important since it could produce an effect on stress and eventually on the perceived pain, as in the case of vaccinations \cite{BERAN2013,Mataric2015}.%,Jeong:2018}. 

The current, not pharmacological, method used by the clinicians to cope with pain is the distraction. Distraction is defined as the use of strategies to take an individual's attention away from the procedure, for example, by asking questions, or by instructing the child to blow. A distraction method by using a robot has been already developed and tested in the case of vaccinations \cite{BERAN2013}. However, the authors did not focused on the role of emotional cues in getting the children attention, and followed the same actions line for all the children, without personalizing the interaction on the base of the child's personal behavior or the emotional state. 

Children distract themselves in different ways and it could be difficult to attract the child's attention when he/she is in a state characterized by and a high level of anxiety due to the incoming medical procedure. Literature on Cognitive Neuroscience is highlighting the influence of emotional related stimuli versus neutral ones in attention and perception. In particular, there is a fundamental role of emotional salience in attention, suggesting that it can facilitate awareness of emotional stimuli in situations where attentional resources are limited \cite{Phelps06}. Moreover, unpleasant emotional stimuli are more rapidly and automatically processed than emotionally neutral stimuli \cite{Phelps06}. Also in psychology literature, several authors have suggested that the attentional system of anxious individuals may be distinctively sensitive to threat-related stimuli \cite{Bar-Haim,Mogg05}, or more generally stimuli with a negative valence. 
% ho aggiunto qui

%Here, we propose to investigate the role of emotional behaviors in capturing the children attention to divert it from painful stimuli.
Some of the results of a two months study (N=69) conducted at the Health-Vaccines Center of the ASL of Terni (Italy) are presented. 
%We considered two different strategies with respect to the initial anxiety level of the child. These two strategies are compared with the baseline where the same cognitive-behavioral distraction strategy was performed by the robot but without any emotional related behavior. According to our knowledge, this is the first attempt to investigate the role of emotional behaviors as a distraction strategy to be used by SAR for pediatric health-care, so we aim at contributing in the effective design of the robot behavior in these settings.

\begin{figure}[!t]
    \centering
\includegraphics[width=0.99\linewidth]{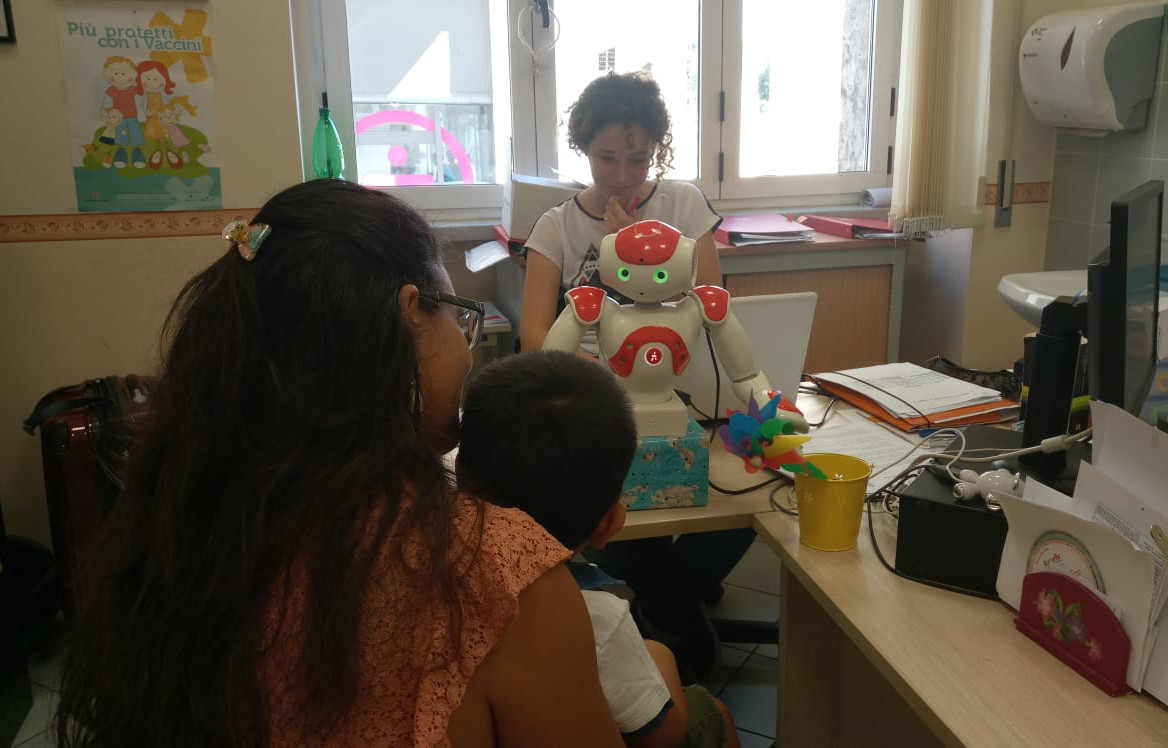}
    \caption{A picture taken during one of the experimental sessions}
    \label{fig:scene}
\end{figure} 

\section{Experimental Design}
Only children aged between 3 to 12 years in the absence of cognitive and/or neurological deficits were included in the study. 
%We considered two different adaptive strategies with respect to the initial anxiety level of the child. These two strategies are compared with a baseline where the same cognitive-behavioral distraction strategy was performed by the robot but without any emotional related behavior. We use the evaluation of the observed reduction in pain and the evaluations of anxiety before, during, and after the vaccination procedure as an index of the distraction ability. 
The robot used for our experimentation is the NAO T14 robot. Single autonomous interacting modules were developed to be activated and sequenced in a WoZ mode in order to be synchronized with the steps of the vaccination procedure. The interaction was organized in three main steps: an introductory and informative part, where the robot asks general questions about the child name and age, providing, in case, procedural and sensory information about the procedure (i), the distraction procedure, where the robot engages the child in talking about general interests such as music and movies (ii) and, finally, a pinwheel blowing task (iii).

The robot was pre-programmed to execute a different interaction strategy, in terms of topics to discuss, with respect to the children age and with two different modulations in terms of nonverbal social cues according to the initial anxiety level of the child. For the emotional behavior, we rely on the use of multiple cues: color (eye LEDs), sound, and motion (robot gestures) \cite{song2017expressing} selected following a valence-arousal classification of the NAO and Pepper behaviors \cite{mina}. 

If the child, before the starting of the procedure, was categorized with a low level of anxiety (AL=1), the robot would select an emotional behavior with a positive value for the valence: the robot would perform happy animation with green colored LEDs, simulating a nice and friendly mood.
On the contrary, if the child was categorized with a high level of anxiety (AL=2), NAO would select emotions with a negative valence following the principle that the attentional system of anxious individuals is more sensitive to those \cite{Bar-Haim,Mogg05}: NAO would act upset, complaining, and showing blue eyes. 
Emotional behaviors were used at the beginning and when switching between the three interaction steps.

Another final test session was made without taking into consideration the kids' AL to experiment the effectiveness of emotional behavior alone (AL=none), so the robot would not exhibit any mood-based animations but only the classic behavior accompanying the natural speech.

%(not used in the control group). The personalization of the behavior were performed on the base of the two input parameters the operator would insert before starting the procedure, i.e. age and anxiety level (AL). 

%Out of the 73 children (27 males and 41 females) which interacted with NAO, only 69 children were considered in the analysis, the other 4, despite the fact that they interacted with the robot, they presented cognitive impairments and for this reason they were excluded.

Questionnaires to assess the children anxiety were given to the parents \textit{before}, \textit{during}, and \textit{after} the vaccination procedure. The provided questions were marked with an alphanumeric code and requested rating values from 0 to 4. Children with a before average anxiety value greater than 2 were assigned to AL=2. %Moreover, the FLACC scale, readjusted to include the considered age group was used. This is a measurement scale used by the operator during the procedure to assess pain for children evaluating respectively the face, the legs, the activity, cry, and consolability of the child on the base of five criteria.

\section{Results}
% DOMANDA: e se mettessimo i risultati ANOVA in maniera più descrittiva? perchè non c’è modo di ignorarli...

%This was a randomized controlled study in which children were randomly assigned to a vaccination session with the robot using emotional behavior (and then successively divided in a group with AL=1 and a group with AL=2) or not (baseline condition with AL=none), while a nurse administered the vaccination. 

Out of the 69 total patients, 68\% showed a low level of anxiety (47 children AL=1) and the 9\% of the sample were found to an high level of anxiety (6 children AL=2). All the children in the last two sessions were assigned to the baseline condition (16 children AL=none).

%ANOVA in maniera descrittiva (cioè eliminando i valori)
Overall, from the ANOVA analysis, it emerged that \textit{anxiety} significantly varied in time during the operation (AL=1 with F[2,138]=10.78 and p$<$.01, AL=2 with F[2,15]=6.77 and p=.008, AL=none with F[2,45]=7.54 and p=.001).
In details, the significant state change was localized between the \textit{before} anxiety state and the \textit{during} state, while it did not changed at the end of the interaction. For the group with AL=1, there is a significant difference also for the couple \textit{during} and \textit{after} (with p=0.03) meaning that the distracting role of NAO showing an emotional behavior, in the case of a low initial anxiety of the children, leads the children to an anxiety level during the interaction (anxiety mean value=0.36) that is even lower than the one after the procedure (anxiety mean value=0.79). This difference is not statistically significant in the case of an initial high anxiety level or in the case of NAO not showing any emotional cue. 

%From such data, we could assume that NAO was able to handle the kids' emotions, independently from their initial AL, reaching an anxiety level with no difference among the groups.

% modifichiamo qui, togliendo i dati...
% Overall, from the ANOVA analysis, it emerged that \textit{anxiety} significantly varied in time during the operation (AL=1 with F[2,138]=10.78 and p$<$.01, AL=2 with F[2,15]=6.77 and p=.008, AL=none with F[2,45]=7.54 and p=.001). In details, the significant state change was localized during the first interaction phase, while it did not changed at the end of the interaction. For the group with AL= 1, there is a significant difference also for the couple \textit{during} (anxiety mean = 0.36)  and \textit{after} (anxiety mean = 0.79) with p=0.03. From such data, we could assume that NAO was able to handle the kids' emotions, independently from their initial AL, reaching an anxiety level with no difference among the groups.
% ...fino a qui

%, as indicated by the parents. Out of the 16 children in the group, 12 resulted in having a low initial anxiety state (AL=none-LOW group) and 4 with an initial anxiety state (AL=none-High).

In order to further evaluate the extent of the emotional strategy with respect the initial anxiety level, we decided to operate an extra classification subdividing also the AL=none group with respect to the initial anxiety state (AL=none-HIGH and AL=none-LOW).
The ANOVA test found significant differences between the \textit{during} values of the anxiety for these two groups, so implying that in the case of a high initial anxiety state the distraction strategy without the emotional cues is not able to attract the child attention leading to a higher during anxiety value. Nevertheless, no statistically significant differences are obtained between the \textit{during} anxiety levels of AL=2 group (anxiety mean value=0.83) and AL=none-HIGH group (anxiety mean value=1.5) due to the few numbers of patients in both the groups. More significant results could have been better highlighted with more data on patients with an initial high level of anxiety (AL=2).

\section{Conclusions}
Results showed that the distraction strategies deployed by the robot were able to reduce the level of anxiety of the child in each experimental group. Moreover, in the absence of the emotional cues, a higher level of reported anxiety was observed during the interaction for children with an high initial anxiety value. Due to the children unpredictable behavior and the rarity of the critical anxiety state occurrence (AL=2), we faced difficulties in collecting enough data regarding this category. In addition, we faced another unintentional issue during the experimentation: the anxiety level, as evaluated before the procedure, can give sometimes a false positive. %That means that, if the child behavior expresses a low level of distress before the procedure starts, at the moment of the procedure, it can rapidly change, and sometimes degenerates to panic. In those cases, NAO would still perform the distraction process as configured for a low level of anxiety, acting, in fact, in the wrong way. This issue highlighted the necessity of providing the robot with a further adaptation ability to be deployed also during the interaction in the case of a change in the anxiety state. 

\section*{Acknowledgements}
This study has been partially supported by MIUR within the PRIN2015 research project ``UPA4SAR - User-centered Profiling and Adaptation for Socially Assistive Robotics'' (grant n. 2015KB-L78T).

\bibliographystyle{IEEEtran}
\bibliography{references}
\end{document}